\documentclass[lettersize,journal]{IEEEtran}
\usepackage{amsmath,amsfonts}
\usepackage{algorithmic}
\usepackage{algorithm}
\usepackage{array}
\usepackage[caption=false,font=normalsize,labelfont=sf,textfont=sf]{subfig}
\usepackage{textcomp}
\usepackage{stfloats}
\usepackage{url}
\usepackage{verbatim}
\usepackage{graphicx}
\usepackage{cite}
\usepackage{multirow}
\usepackage{booktabs}
\usepackage{bbding}
\begin{document}

\title{Forward Consistency Learning with Gated Context Aggregation for Video Anomaly Detection}

\author{Jiahao Lyu, Minghua Zhao, Xuewen Huang, Yifei Chen, Shuangli Du, Jing Hu, Cheng Shi, Zhiyong Lv

\thanks{
	
	Jiahao Lyu, Minghua Zhao, Yifei Chen, Shuangli Du, Jing Hu, Cheng Shi, Zhiyong Lv are with the Shaanxi Key Laboratory for Network Computing and Security Technology, School of Computer Science and Engineering, Xi'an University of Technology, Xi’an, 710048, China. Corresponding author: Minghua Zhao. zhaominghua@xaut.edu.cn

Xuewen Huang is with the School of Cyber Science and Engineering, Xi’an Jiaotong University, Xi’an, 710049, China.
}}


\maketitle

\begin{abstract}

As a crucial element of public security, video anomaly detection (VAD) aims to measure deviations from normal patterns for various events in real-time surveillance systems. However, most existing VAD methods rely on large-scale models to pursue extreme accuracy, limiting their feasibility on resource-limited edge devices. Moreover, mainstream prediction-based VAD detects anomalies using only single-frame future prediction errors, overlooking the richer constraints from longer-term temporal forward information. In this paper, we introduce FoGA, a lightweight VAD model that performs \textbf{Fo}rward consistency learning with \textbf{G}ated context \textbf{A}ggregation, containing about 2M parameters and tailored for potential edge devices. Specifically, we propose a Unet-based method that performs feature extraction on consecutive frames to generate both immediate and forward predictions. Then, we introduce a gated context aggregation module into the skip connections to dynamically fuse encoder and decoder features at the same spatial scale. Finally, the model is jointly optimized with a novel forward consistency loss, and a hybrid anomaly measurement strategy is adopted to integrate errors from both immediate and forward frames for more accurate detection. Extensive experiments demonstrate the effectiveness of the proposed method, which substantially outperforms state-of-the-art competing methods, running up to 155 FPS. Hence, our FoGA achieves an excellent trade-off between performance and the efficiency metric.


\end{abstract}


\section{Introduction}
Video surveillance systems have been widely deployed in the field of public safety, leading to an explosive growth in the volume of recorded video data \cite{liu2024generalized}. However, most surveillance devices merely store video streams and lack the capability to detect anomalous events and issue timely, intelligent alarms. Consequently, the development of video anomaly detection (VAD) techniques that achieve both high accuracy and rapid response has attracted increasing attention from researchers and remains a highly challenging task in the computer vision community.

Anomalies are typically defined as events that deviate from normal patterns or expected behaviors \cite{gong2019memorizing,park2020learning}, characterized by uncertainty and irregularity, such as throwing objects, running, or violating traffic rules. Because of these characteristics, anomalous events occur far less frequently than normal ones in real-world scenarios. Consequently, many anomaly detection datasets are severely imbalanced, with normal events greatly outnumbering anomalous ones \cite{Sun2023Hierarchical,wang2023video}, which poses significant challenges to effective anomaly detection. Thus, performing unsupervised anomaly detection under such highly imbalanced data distributions has become a major research focus \cite{doshi2023towards}.

\begin{figure}[t] 
	\centering
	\includegraphics[width=\linewidth]{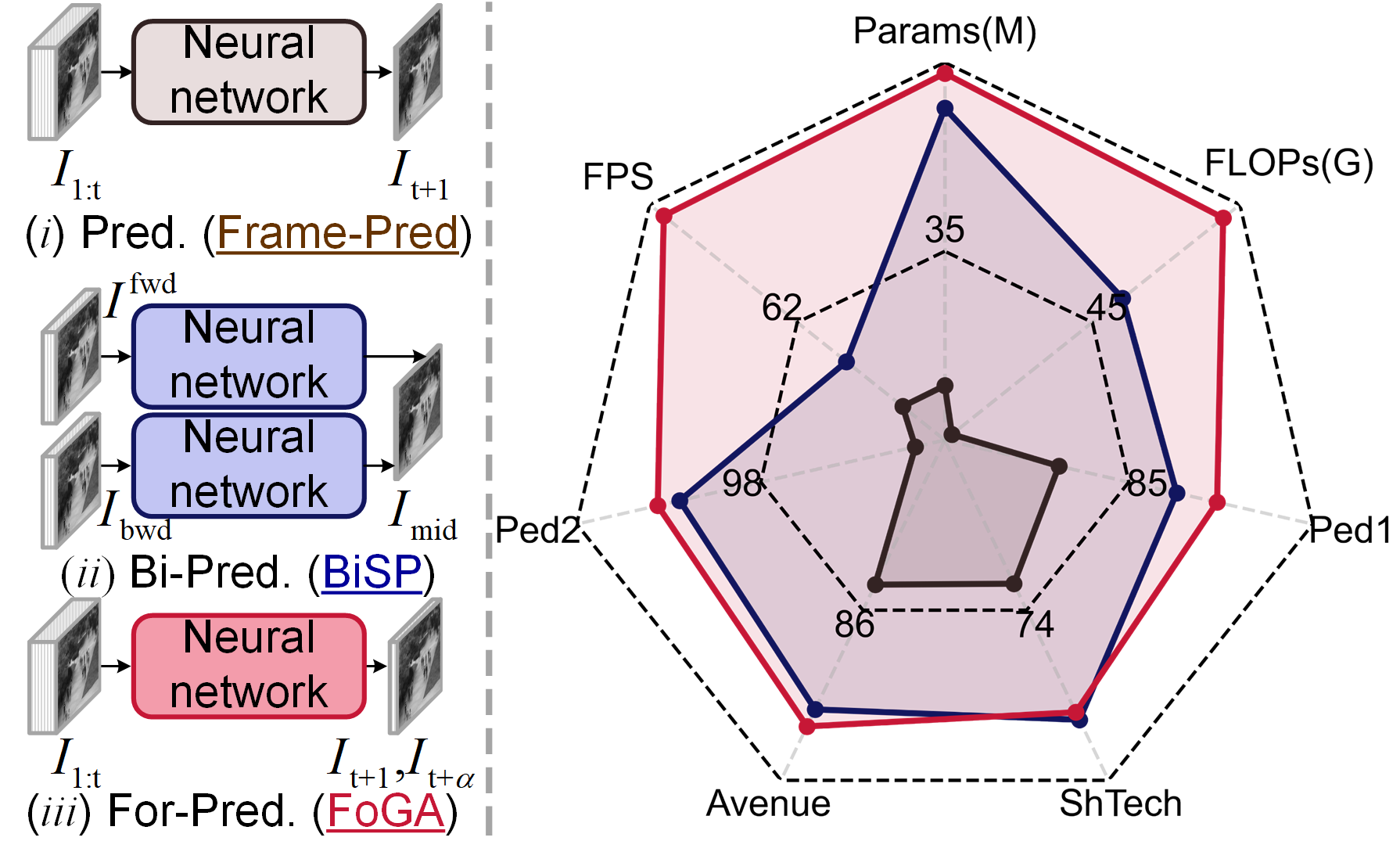}
	\caption{\textbf{\emph{Left:}} Three prediction modeling types of VAD. \emph{(i)} Prediction-based (Pred.) method. \emph{(ii)} Bidirectional prediction-based (Bi-Pred.) method. \emph{(iii)} The proposed forward consistency prediction method (For-Pred.). \textbf{\emph{Right:}} The radar chart provides comparison among Frame-Pred \cite{liu2018future}, BiSP\cite{lyu2026bidirectional} and the proposed FoGA on four datasets performance (AUC $\uparrow$), as well as efficiency metric (FPS $\uparrow$, Params(M) $\downarrow$ and Flops(G) $\downarrow$).} \label{fig1}
\end{figure}

Recently, the unsupervised paradigm has been dominated by reconstruction-based and prediction-based. Reconstruction-based methods \cite{gong2019memorizing,chang2020clustering,zhong2022cascade,kommanduri2024dast,astrid2023pseudobound} first reconstruct the input video frame sequence and then perform anomaly detection based on the error between the reconstructed frames and the original frames. Meanwhile, prediction-based methods \cite{liu2018future,cai2021appearance,hao2022spatiotemporal,zhao2025rethinking,park2025fast,lyu2025moba} predict the next instant frame using the input frame sequence, and detect anomalies by calculating the prediction error between the predicted frame and the ground truth frame. The essence of these two categories of methods resides in extracting spatial or temporal features from video sequences and measuring the degree of deviation of events based on errors, which anomaly events usually produce a larger error than normal.

Compared with prediction-based methods, reconstruction-based methods reconstruct each input frame and primarily model appearance patterns, which inevitably ignores motion information. Consequently, prediction-based VAD methods have been widely explored for their stronger capability to capture temporal dynamics and motion cues. The prediction baseline \cite{liu2018future} employed U-Net \cite{Ronneberger2015Unet} and generative adversarial network (GAN) \cite{goodfellow2014generative} to construct a frame prediction model that generates the next immediate frame. Therefore, most existing methods \cite{hao2022spatiotemporal,huang2022self,yang2023video,zhao2025rethinking,41} only predict the next one. During inference, certain anomalies initially exhibit motion patterns resembling the normal, which makes it difficult to distinguish them accurately and may result in a high false detection rate. Alternatively, as illustrated in Fig.~\ref{fig1}\emph{(ii)}, some works achieve bidirectional prediction \cite{fang2020anomaly,lyu2026bidirectional} by performing both forward and backward prediction, but they are still limited to the prediction of a single frame. Despite various modules being developed by the above methods to improve accuracy, only a few can satisfy the real-time detection requirements of edge-based surveillance systems \cite{park2025fast}. Considering the limitations of edge devices, developing a lightweight VAD model with fewer parameters, lower memory usage, higher accuracy, and real-time performance remains a promising direction.

In summary, Fig.~\ref{fig1}\emph{(iii)} illustrates the proposed FoGA, a lightweight forward consistency prediction method that not only generates immediate frames but also learns to predict longer-term forward frames. As shown in Fig.~\ref{fig1} \textbf{\emph{Right}}, it outperforms existing methods in both AUC and efficiency metrics. To this end, based on the U-net framework, we also design the Gate Context Aggregation Module (GCAM) to dynamically adjust the focus of attention of the model, effectively eliminating the unavoidable redundant information in multi-scale feature extraction. Furthermore, considering that some abnormal and normal behaviors may exhibit similar motion patterns at the beginning of motion, we used the proposed additional forward consistency loss to optimize network training, ensuring that the generated two prediction frames maintain motion consistency. During inference, a multi-scale anomaly scoring strategy is employed to measure the deviations between the two types of predicted frames and their corresponding ground truths, thereby achieving efficient anomaly detection. Our main contributions are as follows:

\begin{itemize} 
	\item{We propose a lightweight VAD that jointly predicts an immediate frame and a longer-term forward frame to strengthen temporal modeling for anomaly detection.}
	\item{We design the Gate Context Aggregation Module within a U-Net backbone to adaptively reweight multi-scale features, filtering redundant responses during feature fusion.}
	\item{We propose an additional forward consistency loss to enhance prediction quality, and employ hybrid errors to measure the deviation of anomalies.}
	\item{Experiments on four benchmarks demonstrate that our method achieves the best trade-off between detection performance and efficiency.}
\end{itemize}

\section{Related Work}

\subsection{Video Anomaly Detection}
\textbf{Unsupervised VAD} aims to learn normal patterns solely from normal samples and to detect anomalies by measuring deviations during inference. Reconstruction-based methods \cite{gong2019memorizing} assume that anomalies cannot be well reconstructed. They compress input frames into low-dimensional features and then reconstruct them, identifying anomalies through reconstruction errors, smaller for normal events and larger for abnormal ones \cite{8,chang2020clustering,morais2019learning,30}. Although GAN-based methods enhance normal reconstruction \cite{24,33,34,zhang2022detecting}, discriminators introduce instability and additional computational costs, and models may still reconstruct abnormal regions similar to normal patterns. Prediction-based methods, on the other hand, aim to forecast future frames using both appearance and motion features. Liu et al. \cite{liu2018future} introduced the first prediction baseline using optical flow for motion optimization and GANs for appearance. Subsequent studies \cite{cai2021appearance,38,39,huang2023video,41} have further improved motion consistency through optical flow, dual-stream network, or multi-branch structures, though at the cost of higher computation. Beyond single-direction prediction, recent works explore bidirectional prediction \cite{zhong2022bidirectional,lyu2026bidirectional} to exploit temporal information from both forward and backward sequences. Methods such as \cite{fang2020anomaly,44} integrate features or consistency losses across directions to enhance motion modeling. However, excessive bidirectional processing increases computational complexity. In contrast, our method employs additional forward consistent frames to efficiently achieve dual-frame prediction in a single direction.

\textbf{Efficient VAD} aims to meet real-time constraints without sacrificing accuracy. While most methods improve detection with large models or multimodal designs, several recent works achieve a better performance--efficiency balance via knowledge distillation~\cite{croitoru2024lightning,deng2023prior} or Mamba-based methods~\cite{lyu2025vadmamba}. Distillation-based methods~\cite{croitoru2024lightning,deng2023prior} transfer anomaly-relevant representations from a strong teacher to a lightweight student for faster inference, whereas VADMamba~\cite{lyu2025vadmamba} jointly models appearance and motion through frame prediction and flow reconstruction with compact normal-pattern representations. However, these methods primarily achieve real-time capability through architectural innovations, yet their designs often rely on complex multi-task optimization and still demand considerable model capacity, which hinders deployment on resource-limited edge devices. Instead, our method employs a single-task, single-model prediction framework with only 2M parameters that enables high real-time power and shows strong potential at the edge.

\begin{figure*}[t]
	\centering
	\includegraphics[width=\textwidth]{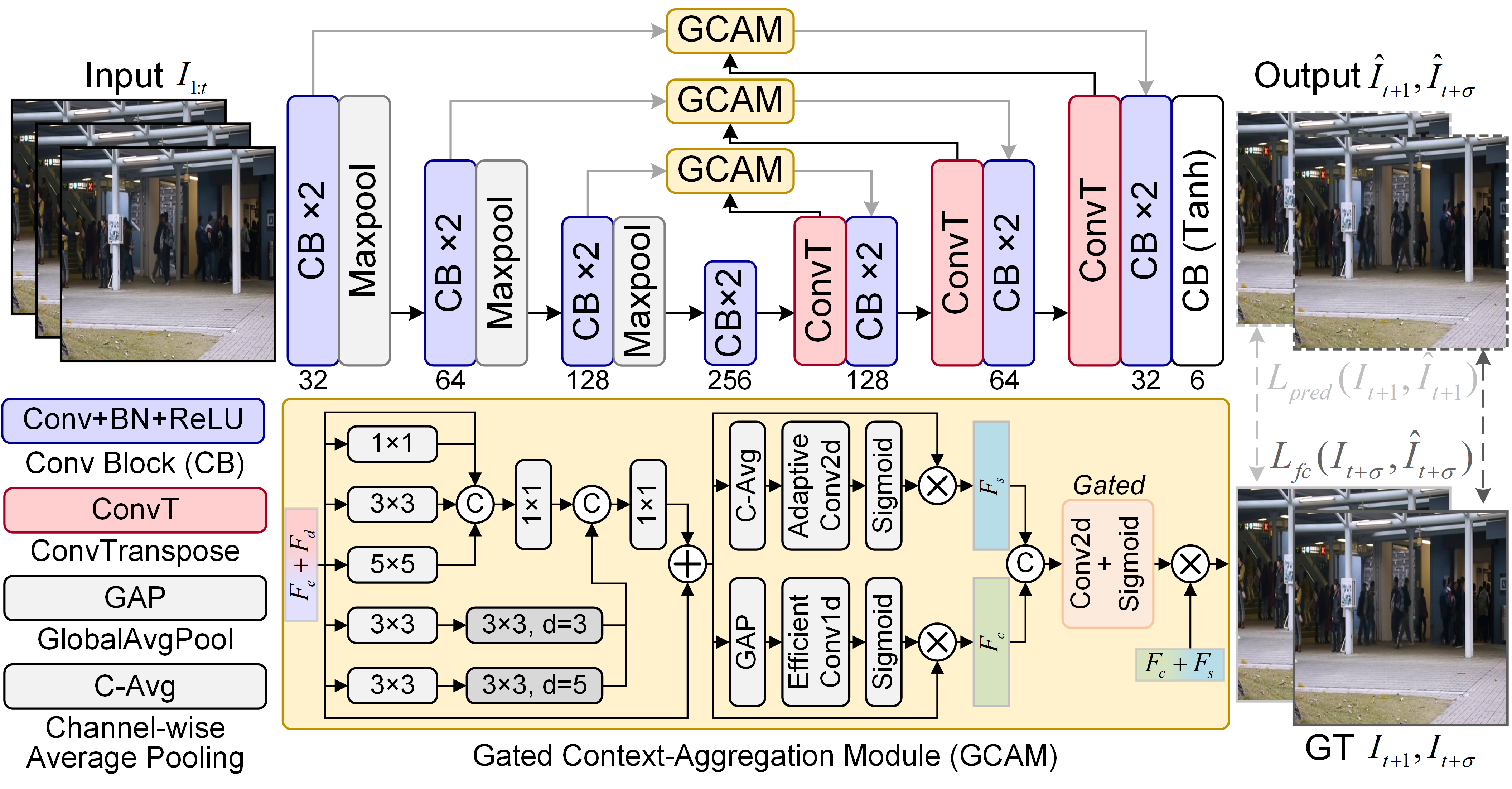}
	\caption{Overall framework of the proposed FoGA that simultaneously predicts an immediate frame $\hat{I}_{t+1}$ and a forward frame $\hat{I}_{t+\sigma}$. It employs a U-Net architecture consisting of the Convolutional Block (CB) and the proposed GCAM. The horizontal axis denotes the number of output channels.}
	\label{fig2}
\end{figure*}

\subsection{Gated Attention}

Gated attention~\cite{qiu2025gated} employs a learned gating mechanism to multiplicatively modulate attention weights, enabling dynamic information routing and richer nonlinear interactions. It also alleviates the computational burden of standard softmax attention in long-range modeling, often achieving sub-quadratic or even linear complexity. In anomaly detection, GT-HAD~\cite{lian2024gt} employs gated self-attention to mitigate identity mapping in reconstruction-based frameworks, thereby enhancing sensitivity to hyperspectral anomalies. Similarly, DBPI~\cite{wang2023dual} integrates a gated-attention unit to selectively fuse normal-appearance priors into the autoencoder, encouraging normal-like reconstruction in abnormal regions and alleviating the industrial anomaly-escape problem. 

Inspired by the success of gated attention in anomaly detection, we extend this mechanism to VAD. We propose a gated context aggregation module that adaptively fuses multi-scale cues for frame prediction, adding only 0.08M parameters each on average.

\section{Methodology}
The overall pipeline of the proposed FoGA is illustrated in Fig. \ref{fig2}, which consists of a lightweight U-Net backbone and a gated context aggregation module (GCAM) embedded into the skip connections. FoGA takes $t$ consecutive frames $I_{1:t}$ as input and predicts multiple forward future frames, $\hat{I}_{t+1}$ and $\hat{I}_{t+\sigma}$. By enforcing forward consistency constraints during training, the model learns more discriminative representations. During inference, FoGA integrates the hybrid errors of different forward frames to achieve more robust anomaly detection performance.

\subsection{Architecture Overview}
As illustrated in Fig.~\ref{fig2}, FoGA adopts an autoencoder as its core backbone to perform forward prediction for video anomaly detection. Given $t$ consecutive input frames $\{I_1, I_2, \dots, I_t\}$, the network models dual temporal dynamics by predicting both the immediate next frame $\hat{I}_{t+1}$ and a longer-range forward frame $\hat{I}_{t+\sigma}$. The design enables the FoGA to capture short-term motion cues as well as long-term timing dependencies. Especially, it can achieve robust detection by the hybrid scores.

The backbone follows a symmetric encoder-decoder architecture. In the encoding stage, consecutive input frames are progressively mapped to high-dimensional representations through Convolutional Blocks (CB) and Maxpool operations, which jointly increase feature dimensionality and reduce spatial resolution to capture rich semantic and contextual information. The decoding stage progressively restores spatial resolution via transposed convolutions (ConvT) followed by convolutional blocks to refine reconstructed feature. The skip connections link corresponding encoding and decoding stages to preserve multi-scale spatial details, within which gated context aggregation is embedded to regulate information flow from encoder to decoder adaptively. Based on the last decoded features, through CB with Tanh to output two predictions, immediate frame $\hat{I}_{t+1}$ and forward frame $\hat{I}_{t+\sigma}$.

\subsection{Gated Context Aggregation Module}
To enhance spatio-temporal representation with minimal overhead, we propose the Gated Context Aggregation Module  (GCAM), a lightweight plug-in that selectively aggregates context features while suppressing redundant responses. GCAM comprises a Context Feature Aggregation module for multi-scale context fusion and an Efficient Gated Attention module that adaptively filters and emphasizes informative features.

\textbf{Context Feature Aggregation Module.} In real-world surveillance scenarios, objects of interest exhibit substantial variations in scale, shape, and spatial distribution across different scenes. Moreover, the same object may appear at different effective scales across feature resolutions during hierarchical representation learning. Relying on a single-scale convolution is therefore insufficient, as it limits feature perception to a fixed receptive field and may bias the model toward dominant regions. As a result, small or large objects, as well as temporal contextual interactions, may be inadequately represented. To address this issue, we propose a context feature aggregation (CFA) module that explicitly captures shallow representations across multiple spatial extents. Specifically, we take the element-wise sum of the encoded feature $\mathbf{F}_e$ and the upsampled feature $\mathbf{F}_d$ as the input to CFA, parallel convolutional branches with kernel sizes of $1\times1$, $3\times3$, and $5\times5$ are employed to model local patterns under diverse receptive fields, enabling the aggregation of both fine-grained and coarse-grained features. Furthermore, to capture potential semantic dependencies among objects and between objects and background regions, dilated convolutions with a $3\times3$ kernel and dilation rates of $3$ and $5$ are incorporated to expand the effective receptive field without additional spatial downsampling. By jointly integrating standard and dilated convolutions across multiple scales, the proposed module enhances contextual perception and alleviates scale-induced bias, resulting in more discriminative and robust feature representations.

\textbf{Efficient Gated Attention Module.} To adaptively regulate information flow and suppress redundant responses in aggregated contextual features, we propose an Efficient Gated Attention (EGA) module that integrates lightweight channel-wise and spatial-wise attention with a gating mechanism. The EGA module enhances feature selectivity while maintaining low computational overhead, making it well suited for real-time anomaly detection.

The channel-wise attention branch adopts an Efficient Channel Attention (ECA) mechanism to model inter-channel dependencies without dimensionality reduction. Given an input feature $\mathbf{X} \in \mathbb{R}^{B \times C \times H \times W}$, global average pooling is first applied to aggregate spatial information and generate channel feature map $\mathbf{f}_c\in\mathbb{R}^{B \times C \times 1 \times 1}$:
\begin{equation}
\mathbf{f}_c = \mathrm{GAP}(\mathbf{X}).
\end{equation}

The pooled feature is then reshaped and processed by a 1D convolution to capture local cross-channel interactions:
\begin{equation}
\mathbf{M_c} = \sigma\!\left(\mathrm{Conv1D}_k(\mathbf{f}_c)\right),
\end{equation}
where $\sigma(\cdot)$ denotes the sigmoid function, and $k$ is an adaptively determined kernel size computed as
\begin{equation}\label{eqk}
k = \left| \frac{\log_2(C) + b}{\gamma} \right|_{\text{odd}},
\end{equation}
where the parameters are set as $b$=1, $\gamma$=2, and $\left| \cdot \right|_{\text{odd}}$ denotes the nearest odd integer. The channel-refined feature map is obtained as:
\begin{equation}
\mathbf{F}_c = \mathbf{X} \odot \mathbf{M_c},
\end{equation}
where $\odot$ denotes element-wise multiplication with channel-wise broadcasting.

To complement channel-wise modulation, the spatial-wise attention branch employs an Efficient Spatial Attention (ESA) module to capture long-range spatial dependencies and generate spatial feature map $\mathbf{f}_s\in\mathbb{R}^{B \times 1 \times H \times W}$. Specifically, the input feature map is first compressed along the channel dimension:
\begin{equation}
\mathbf{f}_s = \mathrm{AvgPool_{c}}(\mathbf{X}).
\end{equation}
where $\mathrm{AvgPool_{c}}$ denotes average pooling across the channel.

Then, using a 2D convolution with an adaptively selected kernel size to generate a spatial attention map $\mathbf{M}_s$.
\begin{equation}
\mathbf{M_s} = \sigma\!\left(\mathrm{Conv2D}_k(\mathbf{f}_s)\right),
\end{equation}
where the kernel size $k$ of is ESA is similar to Eq.(\ref{eqk}), differing in its determination according to spatial resolution:
\begin{equation}
k = \max\!\left(3, \left| \frac{\log_2(HW) + b}{\gamma} \right|_{\text{odd}} \right).
\end{equation}

The spatially refined feature map is given as
\begin{equation}
\mathbf{F}_s = \mathbf{X} \odot \mathbf{M_s}.
\end{equation}

The channel-refined feature $\mathbf{F}_c$ and spatial-refined feature $\mathbf{F}_s$ are concatenated and fed into a gating unit composed of a convolution layer followed by a sigmoid activation to learn adaptive fusion weights:
\begin{equation}
\mathbf{g} = \sigma\!\left(\text{Conv2D}\!\left([\mathbf{F}_c, \mathbf{F}_s]\right)\right).
\end{equation}

Finally, the gated attention output is obtained by modulating the combined features:
\begin{equation}
\mathbf{X}_{\text{out}} = \mathbf{g} \odot (\mathbf{F}_c + \mathbf{F}_s).
\end{equation}

Thereby enabling the EGA module to dynamically balance channel-wise importance and spatial contextual relevance. Through this efficient gated design, the EGA module enhances discriminative feature representations while preserving computational efficiency.


\subsection{Training Loss}

Following \cite{liu2018future,le2023attention}, we employed intensity and gradient functions in the image space to predict $\hat{I}_{t+1}$ and $\hat{I}_{t+\sigma}$. The intensity function constrains the pixel-level similarity between each predicted image and its corresponding ground truth, while the gradient function encourages sharper and more detailed predictions. The intensity function $\mathrm{Int(\cdot)}$ and gradient function $\mathrm{Grad(\cdot)}$ are defined as follows:

\begin{equation}
\mathrm{Int}(\hat{I}_{t}, I_{t})=\left\| {I_{t} - {\hat{I}}_{t}} \right\|_{2}^{2}.
\end{equation}

\begin{equation}
\begin{gathered}
\mathrm{Grad}(\hat{I}_{t}, I_{t})=\sum_{i,j} \left \| \left | I_{t}^{i,j}-I_{t}^{i-1,j} \right | -\left | \hat{I}_{t}^{i,j}-\hat{I}_{t}^{i-1,j} \right | \right \|_{1} \\
+ \left \| \left | I_{t}^{i,j-1}-I_{t}^{i,j} \right | -\left | \hat{I}_{t}^{i,j-1}-\hat{I}_{t}^{i,j} \right | \right \|_{1},
\end{gathered}
\end{equation}
where $I_{t}$ and $\hat{I}_{t}$ represent ground truth frame and predicte frame, respectively, and $i$ and $j$ represent the spatial index of the $t$-th frame.

Based on the above two functions, we propose to the predict loss $\mathcal{L}_{pred}$ and the forward consistency loss $\mathcal{L}_{fc}$ as follows:

\begin{equation}\label{eq1}
\mathcal{L}_{pred}= \text{Int}(\hat{I}_{t+1}, I_{t+1}) + \text{Grad}(\hat{I}_{t+1}, I_{t+1}).
\end{equation} 

\begin{equation}\label{eq2}
\mathcal{L}_{fc}=\text{Int}(\hat{I}_{t+\sigma}, I_{t+\sigma})+ \text{Grad}(\hat{I}_{t+\sigma}, I_{t+\sigma}).
\end{equation}

Furthermore, to improve the structural alignment between the two predicted results and ensure reliable hybrid error quality during inference, we introduce the Structural Similarity Index Measure (SSIM) to define the consistency loss $\mathcal{L}_{con}$:

\begin{equation}\label{eq3}
\mathcal{L}_{con}=1-\operatorname{SSIM}(\hat{I}_{t+1}, \hat{I}_{t+\sigma}).
\end{equation}

Ultimately, the combination of loss function including two prediction losses and consistency loss is shown in Eq.~(\ref{eq4}):

\begin{equation}\label{eq4}
\mathcal{L}=\mathcal{L}_{pred}+\mathcal{L}_{fc}+\mathcal{L}_{con}.
\end{equation}

\subsection{Anomaly Score}

%
%
%

In the inference phase, we employ Peak Signal-to-Noise Ratio (PSNR) to measure the anomaly scores $S(I_{t})$, where PSNR is calculated by the mean-square error $E_{i}(I_{t}, \hat{I}_{t})$ between the ground truth $I_{t}$ and the predicted frame $\hat{I}_{t}$. Since the forward consistency constraint of FoGA, we thus fused the immediate $E_{i}(I_{t+1}, \hat{I}_{t+1})$ and forward $E_{f}(I_{t+\sigma}, \hat{I}_{t+\sigma})$ errors by the weighted sum strategy, as shown in Eq.~(\ref{eq6.1}):

\begin{equation}\label{eq6.1}
E=E_{i} + \lambda \cdot E_{f},
\end{equation}
where $\lambda$ is the weights of the $E_{f}$.

Based on the hybrid error $E$, we adopt the multi-scale anomaly evaluation method \cite{zhong2022bidirectional} to compute the anomaly score. This measurement method achieves more comprehensive anomaly detection across three scales by constructing an error pyramid, as formulated in Eq.~(\ref{eq5}).

\begin{equation}\label{eq5}
\operatorname{PSNR}(I_{t}, \hat{I}_{t})=10 \log _{10}(\frac{1}{\sum_{i=0}^{N=3} v_{i}}), 
\end{equation}
where $v_{i}$ denotes the maximum prediction mean-square error of the patch block at scale $i$, obtained through mean pooling. Finally, the PSNR is normalized to the range $[0,1]$ using Eq.~(\ref{eq6}) and further smoothed with a 1D Gaussian filter to generate the final anomaly score.

\begin{equation}\label{eq6}
S(I_{t})=\frac{\operatorname{PSNR}(I_{t}, \hat{I}_{t})-\min (\operatorname{PSNR}(I_{t}, \hat{I}_{t}))}{\max (\operatorname{PSNR}(I_{t}, \hat{I}_{t}))-\min(\operatorname{PSNR}(I_{t}, \hat{I}_{t}))}.
\end{equation}

\section{Experiments}

\subsection{Datasets}
In order to evaluate the effectiveness of the proposed method, a number of experiments and ablation studies have been done in this paper on four challenging datasets, i.e., UCSD Ped1 \& Ped2, CUHK Avenue, and Shanghaitech. 

\begin{itemize}
	\item \textbf{UCSD Ped1.} The Ped1 \cite{Mahadevan2010Anomaly} dataset consists of 70 video clips, containing 34 train videos and 36 test videos, with a total of 40 abnormal events, such as people riding bicycles, people crossing the aisle, and some other abnormal events.
	
	\item \textbf{UCSD Ped2.} The Ped2 \cite{Mahadevan2010Anomaly} dataset consists of 28 video clips, containing 16 train videos, 12 test videos, and a total of 12 abnormal events. The anomaly event definition of Ped2 is the same as Ped1, and the abnormal events are mainly caused by vehicles such as bicycles, cars, and so on.
	
	\item \textbf{CUHK Avenue.} The Avenue \cite{lu2013abnormal} dataset consists of 37 video clips containing 16 train videos, 21 test videos, and a total of 47 abnormal events, such as throwing objects, hanging around, walking in the wrong way, and other abnormal events. These video clips have variety in terms of capture mode, lighting, viewing angle, and scale of the abnormal objects. 
	
	\item \textbf{ShanghaiTech.} The Sh-Tech \cite{luo2017revisit} dataset consists of 437 video clips containing 330 train videos, 107 test videos, and a total of 130 anomaly events. With complex lighting conditions and camera angles, rich-scale variations of anomaly objects, large data volumes, and rich scenes, it is a very challenging anomaly detection dataset.
\end{itemize}

\subsection{Experimental Setup}

We implement FoGA using PyTorch and train it with the Adam optimizer at an initialized learning rate of $2e-4$. The input frames are resized to $ 224 \times 224 $, and the pixel values are normalized to $[-1, 1]$. Except for the input frame length $t$ of Avenue, which is set to 8, all other datasets use $t=4$, and the forward prediction parameter $\sigma$ is also set to 4 across all datasets. The error fusing weights $\lambda$ for Ped1, Ped2, Avenue, and Sh-Tech are set to 0.06, 1.0, 0.2, and 0.06, respectively.

\subsection{Quantitative Comparison With Existing Methods}

We compare the frame-level AUC and inference speed (FPS) of FoGA with various state-of-the-art methods under different paradigms in Table~\ref{tab:Comparison}, including 6 reconstruction-based (Recon.) methods~\cite{gong2019memorizing,park2020learning,chang2020clustering,zhong2022cascade,astrid2023pseudobound,kommanduri2024dast}, 14 single-frame prediction-based (Single-Pred.) methods~\cite{liu2018future,park2020learning,hao2022spatiotemporal,huang2022self,yang2022dynamic,zhang2022hybrid,cheng2023spatial,le2023attention,yang2023video,liu2023msn,huang2024long,park2025fast,zhao2025rethinking,lyu2025moba}, and 4 multi-frame prediction-based (Multi-Pred.) methods~\cite{fang2020anomaly,zhong2022bidirectional,li2023multi,lyu2026bidirectional}.

As shown in Table~\ref{tab:Comparison}, our method consistently surpasses most existing reconstruction- and prediction-based baselines, achieving 87.4\%, 98.9\%, 90.1\%, and 76.2\% on Ped1, Ped2, Avenue, and Sh-Tech, respectively. Reconstruction-based methods mainly focus on appearance recovery and are less effective at modeling temporal dynamics. In contrast, single-frame prediction methods exploit motion cues and generally achieve higher detection accuracy, though their performance–efficiency balance varies widely across architectures. Although FastAno++ \cite{park2025fast} attains 130~FPS, its accuracy remains limited. Moreover, existing multi-frame prediction methods often adopt bidirectional prediction using two separate networks, leading to higher computational and memory costs. In contrast, FoGA employs a single autoencoder and achieves state-of-the-art accuracy across benchmarks while maintaining real-time 110~FPS, demonstrating a favorable performance–efficiency trade-off.

\begin{table}
	\centering
	\caption{Comparison with state-of-the-art methods on four benchmark datasets. The best and second-best performances are marked in \textbf{bold} and \underline{underlined}. The comparison methods are listed in chronological order.}
	\label{tab:Comparison}
	\setlength{\tabcolsep}{1.01mm}{
		\begin{tabular}{clcccccc}
			\toprule
			& Method & Reference & Ped1 & Ped2 & Avenue & Sh-Tech & FPS \\
			\midrule
			\multirow{6}{*}{\rotatebox{90}{Recon.}} 
			& MemAE \cite{gong2019memorizing}  & ICCV19 & – & 94.1 & 83.3 & 71.2 & 31\\
			& MNAD \cite{park2020learning}  & CVPR20& – & 90.2 & 82.8 & 69.8 & –\\
			& Cluster-AE \cite{chang2020clustering} & ECCV20& – & 96.5 & 86.0 & 73.3 &– \\
			& CascadeRecon \cite{zhong2022cascade} &PR22 & 82.6 & 97.7 & 88.9 & 70.7 &– \\
			
			& PseudoBound \cite{astrid2023pseudobound} & NC23 & –& 98.4 & 87.1   & 73.7 & –\\
			& DAST-Net \cite{kommanduri2024dast}  &NC24 & 85.4 & 97.9 & 89.8 & 73.7& 22 \\
			\midrule
			
			\multirow{14}{*}{\rotatebox{90}{Single-Pred.}}
			& Frame-Pred \cite{liu2018future} & CVPR18& 83.1 & 95.4 & 85.1 & 72.8 & 25\\
			& MNAD \cite{park2020learning}  & CVPR20& 81.1 & 97.0 & 88.5 & 70.5& 87 \\
			& STCEN \cite{hao2022spatiotemporal} &  ECCV22  & 82.5 & 96.9 & 86.6 & 73.8&  40\\ 
			& SSAGAN \cite{huang2022self} &TNNLS22 & 84.2 & 96.9 & 88.8 & 74.3 & 40 \\
			& DLAN-AC \cite{yang2022dynamic} &  ECCV22 & – & 97.6 & 89.9 & 74.7 & –\\
			& HAMC \cite{zhang2022hybrid} & TCSVT23 & 85.2 & 95.8 & 84.9 & 71.4 & –\\
			& STGCN-FFP \cite{cheng2023spatial} & ICASSP23 & – & 96.9 & 88.4 & 73.7 & –\\
			& ASTNet \cite{le2023attention} &AI23 & – & 97.4 & 86.7 & 73.6&  –\\
			& USTN-DSC  \cite{yang2023video} &  CVPR23 & –   & 98.1 & \underline{89.9}   & 73.8& –\\ 
			& MSN-net \cite{liu2023msn}  & ICASSP23  & – & 97.6 & 89.4 & 73.4 & 95\\
			& PDM-Net \cite{huang2024long} &IJCAI24 & 85.2 & 97.7 & 88.1 & 74.2 & –\\
			& FastAno++ \cite{park2025fast} &PR24 & – &	98.1 & 87.8 & 75.2  & \textbf{130} \\
			& LGN-Net \cite{zhao2025rethinking} &ESWA25 & – & 97.1 & 89.3   & 73.0 & 19 \\
			& MoBA-flow \cite{lyu2025moba} &  KBS26  & – & 98.4 & 88.7   & 75.6 & 68 \\
			\midrule 
			\multirow{5}{*}{\rotatebox{90}{Multi-Pred.}}
			
			& SIGnet \cite{fang2020anomaly}&TNNLS22 & 86.0 & 96.2 & 86.8 & –&  –\\
			& DEDDnet \cite{zhong2022bidirectional} & TCSVT23& – & 98.1 & 89.0 & 74.5 &– \\
			& MGAN-CL \cite{li2023multi} & TCSVT23 & – & 96.5 & 87.1 & 73.6 &  30 \\
			& BiSP \cite{lyu2026bidirectional}&PR26& \underline{86.3} & \underline{98.6} & 89.5 & \textbf{76.4}& 45 \\
			& FoGA & &\textbf{87.4} & \textbf{98.9} & \textbf{90.1}   & \underline{76.2} & \underline{110} \\ 
			\bottomrule
		\end{tabular}
	}
\end{table}

As shown in Fig.~\ref{fig1} \textbf{\emph{Right}}, the radar chart provides an intuitive but partial visualization of efficiency across methods. To present a more comprehensive comparison, Table~\ref{tab:fps} compares parameters count (Params), FLOPs, peak GPU memory (Mem), and FPS across representative baselines~\cite{gong2019memorizing,park2020learning,liu2018future,lyu2026bidirectional} and our FoGA. Except for Frame-Pred~\cite{liu2018future}, all baselines are re-implemented and evaluated on the same GPU. Memory-based methods \cite{gong2019memorizing,park2020learning} introduce additional runtime memory cost, and MemAE~\cite{gong2019memorizing} in particular incurs substantially higher FLOPs due to its 3D convolutional operations. Bidirectional-based method \cite{lyu2026bidirectional} requires dual network inference, which further increases computation and memory usage. FoGA uses the fewest parameters and FLOPs with the lowest peak memory, while achieving the fastest inference, reaching 155 FPS when multi-scale evaluation is disabled.

\begin{table}[h]
	\centering
	\caption{Comparison of model complexity and inference speed. 'M' denotes million. 'G' denotes billion. $^\ast$ denotes re-implemented on the same GPU. () denotes without multi-scale evaluation.}
	\label{tab:fps}
	\begin{tabular}{lcccc}
		\toprule
		Method & Params(M) & Flops(G) & Mem(G) & FPS \\
		\midrule
		Frame-Pred \cite{liu2018future}  & 59.9      & 87.8 & -    &  25 \\
		MemAE \cite{gong2019memorizing}  & 6.50     & 149.04 & 0.79 & 31$^\ast$    \\
		MNAD  \cite{park2020learning}    & 15.00     & 57.49 & 1.02 & 87$^\ast$    \\
		BiSP \cite{lyu2026bidirectional} & 8.45      & 35.85 & 1.25   & 45 (75)$^\ast$  \\
		FoGA                             & \textbf{2.17}      & \textbf{5.85}  & \textbf{0.28} & \textbf{110 (155)}  \\
		\bottomrule
	\end{tabular}
\end{table}

\subsection{Ablation Study}

To assess the influence of different components and hyperparameters on anomaly detection, we conduct a series of ablation studies, including an analysis of the anomaly fusion parameter $\lambda$, an ablation study on the forward constraint parameter $\sigma$, an evaluation of the attention module, and an examination of the impact of removing individual loss terms. Notably, the optimal $\lambda$ selected from the first study is used in the remaining three ablation experiments.

\begin{figure}[h]
	\centering
	\includegraphics[width=\linewidth]{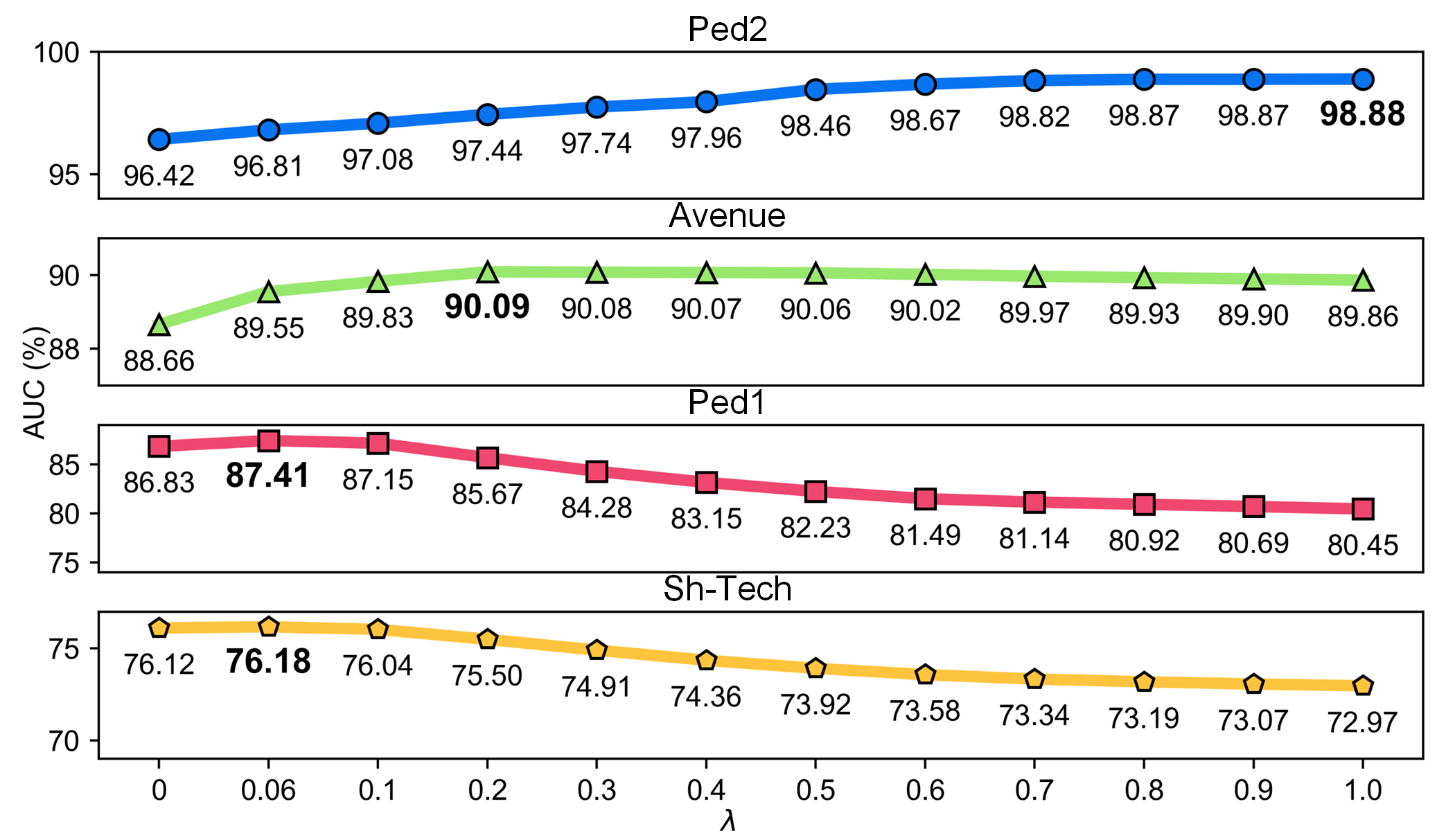}
	\caption{AUC comparison under different $\lambda$ across four datasets.}
	\label{fig3}
\end{figure}

The ablation results in Fig.~\ref{fig3} indicate that the effect of $\lambda$ varies across datasets. Ped2 benefits from relatively larger $\lambda$ values, showing improved regularization stability, whereas Ped1, Avenue, and ShanghaiTech perform better with smaller $\lambda$, suggesting that lighter regularization helps preserve discriminative motion cues. Overall, a modest $\lambda$ offers a balanced performance across all benchmarks.

\begin{figure}[h]
	\centering
	\includegraphics[width=\linewidth]{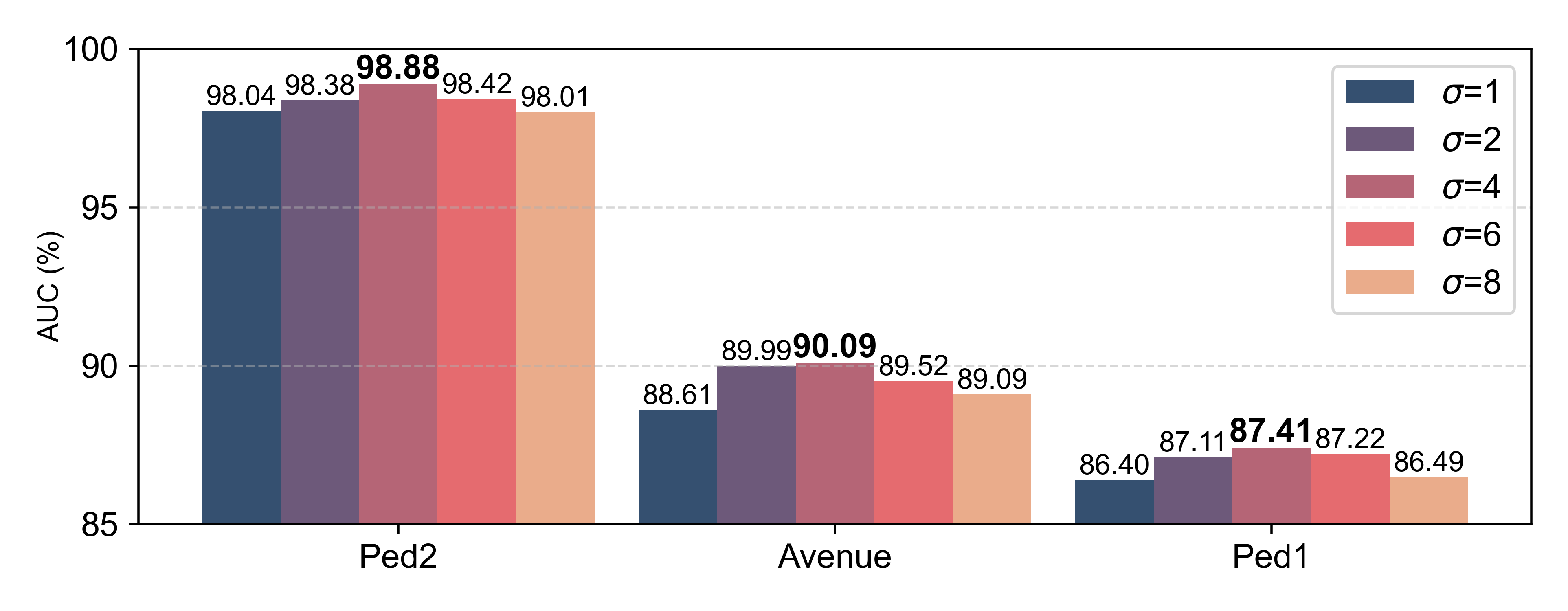}
	\caption{Effect of different forward consistency measurement parameter $\sigma$ across different datasets.}
	\label{fig3.1}
\end{figure}

Fig.~\ref{fig3.1} illustrates the effect of the forward constraint measurement parameter $\sigma$. the FOGA predicts two future frames at the time points $t+1$ and $t+\sigma$, and the forward consistency term measures their corresponding deviation. When $\sigma=1$, the two predictions are consistent, leading to constraint failure, which produces the lowest AUC. For $\sigma>1$, the constraint becomes valid by measuring the time evolution at different scales, which improves the performance, where the appropriate $\sigma$ yields the best results.

Table~\ref{tab:ABLATION} presents the ablation results of CFA and EGA with respect to model complexity and detection accuracy. Activating CFA leads to consistent yet modest improvements with a noticeable computational overhead (+0.20M parameters, +1.47G FLOPs), whereas enabling EGA offers greater performance gains at only a minor cost (+0.04M, +0.31G), demonstrating a more favorable efficiency–accuracy balance. When both modules are applied, the model attains the best overall performance, indicating their complementary roles, though accompanied by the highest complexity (+0.24M, +1.79G), which reflects the expected trade-off of combining both attention modules.

\begin{table}[h]
	\centering
	\caption{Component ablation study of the proposed model.}
	\label{tab:ABLATION}
	\begin{tabular}{ccccccc}
		\toprule
		CFA&EGA&Params(M)& Flops(G) & Ped1 & Ped2& Avenue  \\
		
		\midrule
		\XSolidBrush &\XSolidBrush   & 1.93         & 4.06        & 85.1& 97.2 & 87.3 \\
		\Checkmark & \XSolidBrush  	 & 2.13 (+0.20) & 5.53 (+1.47)& 86.8& 97.4& 88.5\\
		\XSolidBrush &\Checkmark   	 & 1.97 (+0.04) & 4.37 (+0.31)& 86.4& 98.1& 89.0\\
		\Checkmark&  \Checkmark      & 2.17 (+0.24) & 5.85 (+1.79)& 87.4& 98.9& 90.1\\
		\bottomrule
	\end{tabular}
\end{table}

\begin{figure*}[t]
	\centering
	\includegraphics[width=\linewidth]{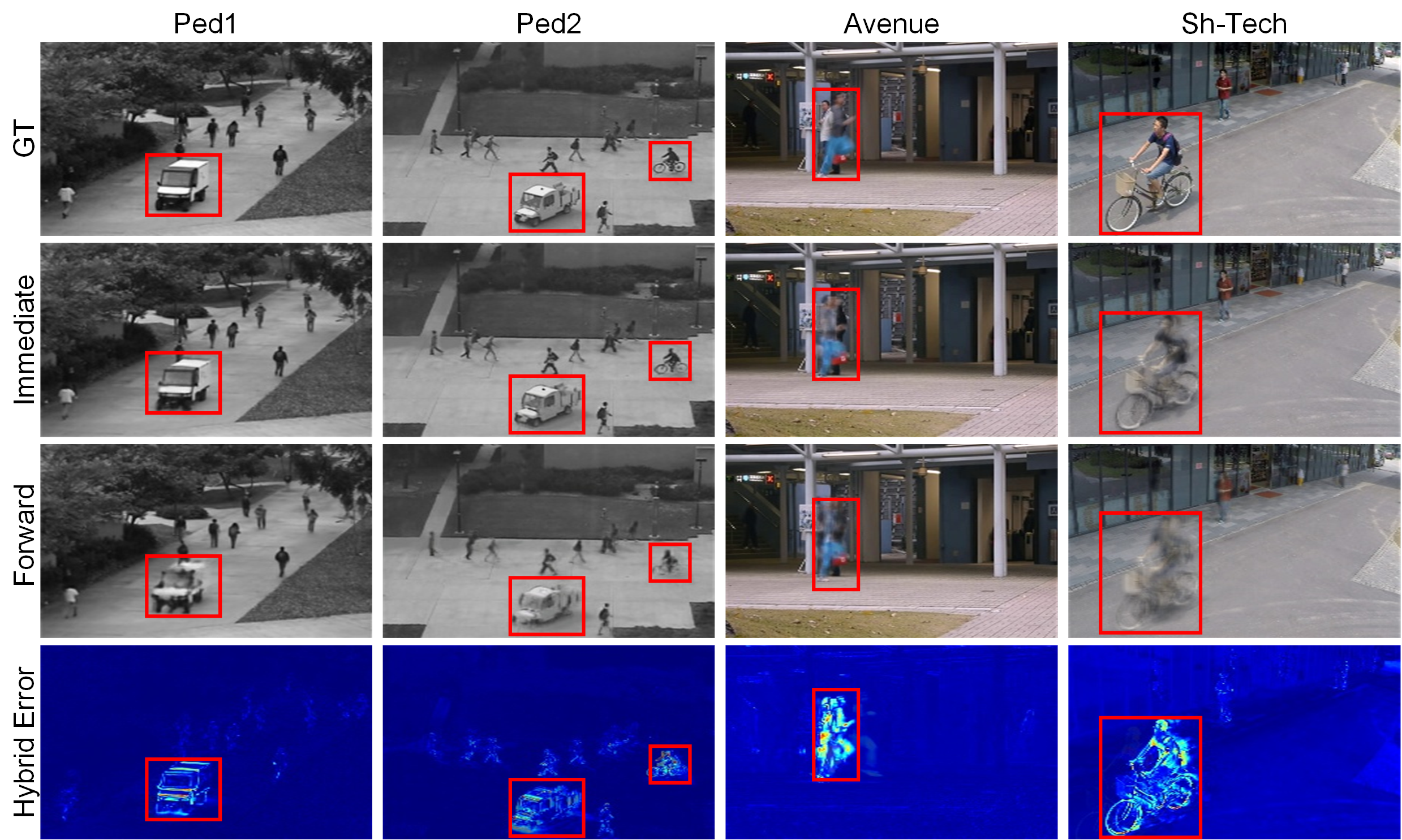}
	\caption{Visualization of Ground Truth (GT) frames, Immediate predictes, Forward predictes, and Hybrid errors on four datasets. The objects marked with red borders are the anomaly. The brighter colors denote larger values of hybrid error.}
	\label{fig4}
\end{figure*}

Table~\ref{tab:loss} presents a loss ablation on Ped1 and Ped2. The exclusion of $\mathrm{Grad}(\cdot)$ results in a noticeable performance decline, highlighting the importance of gradient-based constraints, particularly for motion- and edge-sensitive scenes. Among all losses, $\mathcal{L}_{pred}$ is the most critical supervision, as its removal causes the largest degradation, confirming that it is the cornerstone of FoGA and the core loss for stable training. Both $\mathcal{L}_{fc}$ and $\mathcal{L}_{con}$ further improve the results, with $\mathcal{L}_{fc}$ providing the most consistent gain, which supports the effectiveness of enforcing forward consistency in strengthening temporal modeling to assist anomaly detection effectively. We do not ablate the intensity term $\mathrm{Int}(\cdot)$, because it serves as the fundamental pixel-level supervision. Without it, training becomes difficult to converge, and the anomaly scoring becomes unreliable.

\begin{table}[h]
	\centering
	\caption{Component ablation study of the loss function.}
	\label{tab:loss}
	\begin{tabular}{ccccc}
		\toprule
		\emph{w/o} 	 & $\mathrm{Grad(\cdot)}$	& $\mathcal{L}_{pred}$ & $\mathcal{L}_{fc}$	 & $\mathcal{L}_{con}$ \\
		\midrule
		Ped1	 & 85.5  & 85.0  &  86.9  & 86.4 \\
		Ped2	 & 97.4  & 96.8  &  98.2  & 97.8 \\
		\bottomrule
	\end{tabular}
\end{table}

\subsection{Visualization Analysis}

To further illustrate the forward consistency constraint in FoGA, Fig.~\ref{fig4} presents several prediction results and hybrid error maps on Ped1, Ped2, Avenue, and Sh-Tech. The error maps are computed by taking the difference between each predicted frame and its ground truth, where brighter regions correspond to larger hybrid errors. Comparing the two predictions, the errors in normal regions remain consistently small, often close to zero, indicating that the model can stably reproduce normal appearance and motion patterns. In contrast, the predictions over abnormal regions become visibly blurred, and this degradation is more pronounced for the forward prediction $\hat{I}_{t+\sigma}$ than for the immediate prediction $\hat{I}_{t+1}$. The larger errors at a longer prediction horizon suggest that abnormal behaviors lead to increasing deviations when the model is trained to follow consistent motion evolution, which supports the effectiveness of the proposed forward consistency design in FoGA.

\begin{figure}[h]
	\centering
	\includegraphics[width=\linewidth]{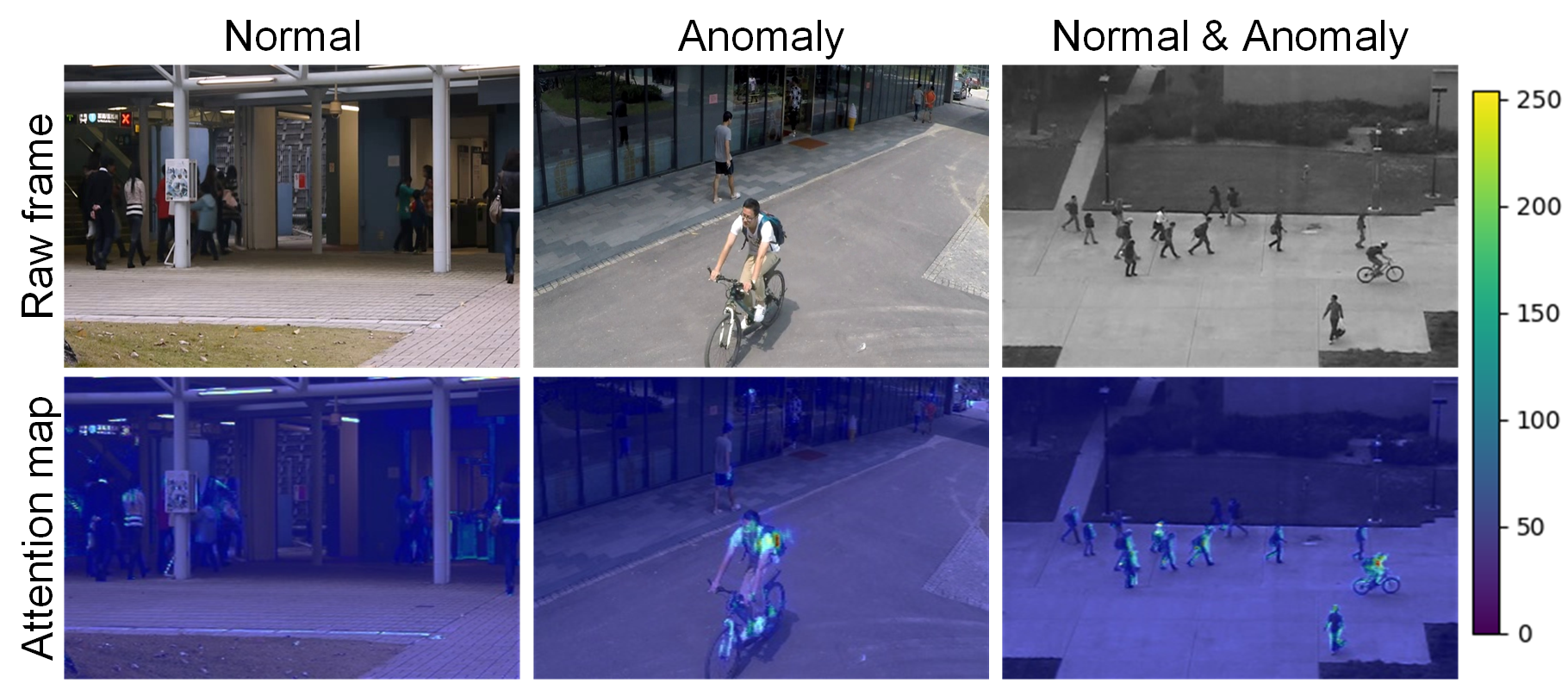}
	\caption{Visualization of Attention map in three different types. The top presents the original frames, and the bottom displays the corresponding attention maps, where brighter regions represent higher attention intensity.}
	\label{fig5}
\end{figure}

Furthermore, fig.~\ref{fig5} visualizes attention maps in three scenarios: Normal, Anomaly, and Mixed (Normal \& Anomaly). In the normal scenario, attention remains weak and diffuse, primarily covering routine activity areas without emphasizing specific targets. In the anomaly scenario, attention becomes sharply concentrated on abnormal objects and motion-related regions (e.g., salient contours and moving parts), while background activations are largely suppressed. In the mixed scenario, the model consistently focuses on the anomalous subject among multiple normal pedestrians, maintaining low responses for regular behaviors. These observations demonstrate that the proposed attention mechanism dynamically enhances focus on relevant motion cues while suppressing distractions, leading to more precise localization of anomalies across diverse scenes.

\begin{figure}[h]
	\centering
	\includegraphics[width=\linewidth]{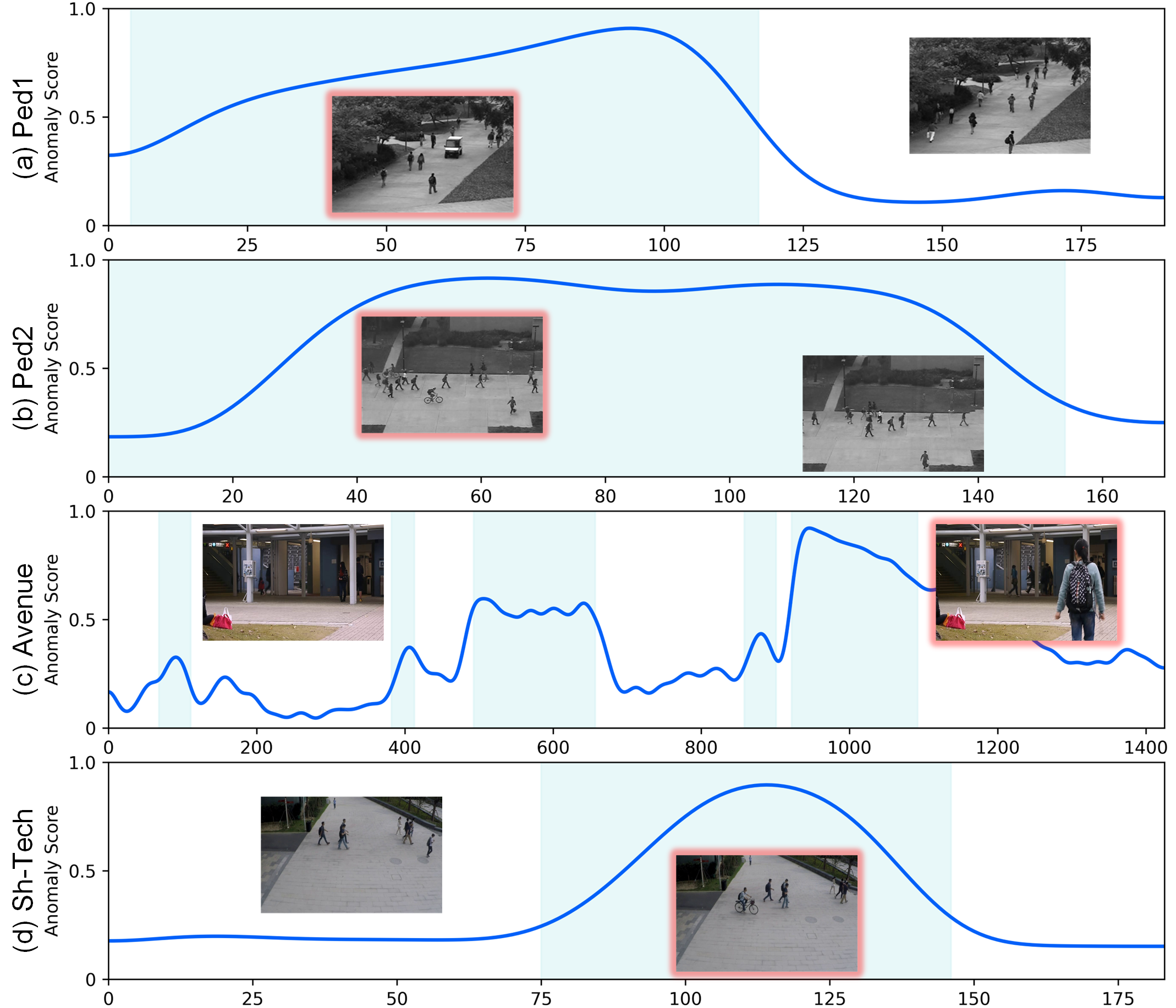}
	\caption{Visualization of score curves on four datasets. The x-axis indicates the video frame index, and the y-axis denotes the anomaly score. Blue blocks mark the Ground-Truth anomalous clips, and the red regions highlight frames containing abnormal events.}
	\label{fig6}
\end{figure}

Fig.~\ref{fig6} shows the frame-level anomaly score curves on Ped1, Ped2, Avenue, and Sh-Tech. The x-axis indicates the frame index, and the y-axis denotes the normalized anomaly score. The blue shaded regions correspond to the ground-truth anomalous intervals, and the red highlights show representative abnormal frames from these segments. Across all datasets, the scores remain low and stable on normal frames, then increase markedly within the annotated abnormal intervals, often reaching a clear peak before returning to a normal level, which indicates accurate temporal localization and good separation between normal and abnormal events. Avenue presents multiple fluctuating peaks, matching its more scattered and longer-lasting anomalies, while Ped1, Ped2, and Sh-Tech show more compact and concentrated responses around the abnormal periods.

\section{Conclusion}

In this paper, we presented FoGA, an efficient and lightweight prediction-based method for VAD that strengthens temporal modeling by learning both immediate and forward-frame predictions. To improve feature utilization with minimal overhead, we introduced a GCAM that adaptively fuses multi-scale cues during prediction. Moreover, we designed a forward consistency loss to align motion evolution across a long-term prediction, together with a hybrid anomaly scoring measurement that better captures deviations in both appearance and dynamics. Extensive experiments on four benchmarks demonstrate that FoGA achieves state-of-the-art detection accuracy while maintaining real-time inference with a compact model size, and ablation studies verify the effectiveness of each component. In future work, we will explore extending FoGA to more diverse open-world scenarios for practical edge deployment.

\section*{Acknowledgments}
This work was supported by the Natural Science Foundation of Shaanxi Province, China (2024JC-ZDXM-35), National Natural Science Foundation of China (No.62571425), and Doctoral Dissertation Innovation Fund of Xi’an University of Technology (BC202621).

\bibliographystyle{IEEEtran}
\bibliography{references.bib}

\newpage

%
%
%
%

\vfill

\end{document}